# Real-Time Object Detection and Classification using YOLO for Edge FPGAs

Rashed Al Amin
*Institute for Embedded Systems*
*University of Siegen*
Siegen, Germany
rashed.amin@uni-siegen.de

Roman Obermaisser
*Institute for Embedded Systems*
*University of Siegen*
Siegen, Germany
roman.obermaisser@uni-siegen.de

*Abstract*—Object detection and classification are crucial tasks across various application domains, particularly in the development of safe and reliable Advanced Driver Assistance Systems (ADAS). Existing deep learning-based methods such as Convolutional Neural Networks (CNNs), Single Shot Detectors (SSDs), and You Only Look Once (YOLO) have demonstrated high performance in terms of accuracy and computational speed when deployed on Field-Programmable Gate Arrays (FPGAs). However, despite these advances, state-of-the-art YOLO-based object detection and classification systems continue to face challenges in achieving resource efficiency suitable for edge FPGA platforms. To address this limitation, this paper presents a resource-efficient real-time object detection and classification system based on YOLOv5 optimized for FPGA deployment. The proposed system is trained on the COCO and GTSRD datasets and implemented on the Xilinx Kria KV260 FPGA board. Experimental results demonstrate a classification accuracy of 99%, with a power consumption of 3.5W and a processing speed of 9 frames per second (FPS). These findings highlight the effectiveness of the proposed approach in enabling real-time, resource-efficient object detection and classification for edge computing applications.

## I. INTRODUCTION

The rapid advancement of computer vision technologies has led to a growing demand for robust and efficient object detection and classification systems across various application domains, including autonomous driving, healthcare, surveillance, robotics, and smart cities. This increasing demand has, in turn, stimulated significant research interest from both academia and industry. Object detection and classification remain challenging, especially in real-time scenarios with critical computational and energy constraints. Various deep learning-based approaches have been developed to improve accuracy, robustness, and speed. Among the prominent methods in this area are CNN and SSD, which have shown better performance in terms of classification accuracy. However, both approaches face limitations in detecting small or densely packed objects [1], which are often encountered in real-world scenes. In contrast, the YOLO family of algorithms has demonstrated high performance in real-time applications, offering a good balance between speed and accuracy [2]. Despite their advantages, these deep learning models typically require substantial computational power and high energy consumption, which restricts their deployment in edge devices and resource-constrained environments.

To overcome these limitations, FPGAs have emerged as a promising hardware platform for accelerating deep learning inference tasks. FPGAs offer re-programmability, high parallelism, and energy efficiency, making them suitable for real-time object detection and classification [3]. They are increasingly adopted in a variety of domains, including advanced driver assistance systems (ADAS) [4], medical informatics [5], traffic light classification [6], waveform generation [7], structural health monitoring [8], super-resolution imaging [9], and general object detection applications [10, 11]. These capabilities have prompted the integration of YOLO-based models with FPGA platforms to leverage their combined strengths. Nevertheless, achieving resource efficiency and maintaining high performance remain key research challenges, particularly when deploying such systems on edge FPGAs with limited hardware resources.

This study proposes a resource-efficient real-time object detection and classification system that integrates the YOLO v5 architecture with FPGA hardware, targeting edge applications. The proposed system leverages the Common Objects in Context (COCO) dataset for general-purpose object detection and the German Traffic Sign Recognition Benchmark (GTSRB) dataset for specialized traffic signal detection and classification. These datasets ensure that the model can perform broad and domain-specific tasks with high accuracy. The system is implemented and evaluated on the Xilinx Kria KV260 FPGA board, a representative edge computing platform. Experimental results demonstrate that the proposed system achieves high classification accuracy while maintaining low power consumption and satisfactory computational speed, highlighting its suitability for real-time deployment in resource-constrained environments. The key contributions of this paper are summarized as follows:

1. Developing a resource-efficient object detection and classification system using YOLO v5 for FPGAs.

2. Optimization of the YOLO model to ensure compatibility with edge FPGA platforms.

3. Performance evaluation and comparison of the proposed system against state-of-the-art FPGA-based object detection and classification systems.

## II. RELATED WORK

Ngo et al. [12] present a real-time traffic sign recognition system implemented on the Xilinx Kria KV260 FPGA platform using the YOLOv3-Tiny network. The system has been designed to address the limitations of GPU-based approaches, such as high power consumption and inefficiency in mobile environments, by leveraging the Deep Learning Processing Unit (DPU) on the FPGA. The authors trained the YOLOv3-Tiny model using a custom VN-ICC dataset of 40 Vietnamese traffic sign classes, achieving a mean Average Precision (mAP) of 94.06% in software and 91.87% on hardware with up to 29 FPS. The implementation process includes model training, conversion to TensorFlow format, quantization to 8-bit integers for performance optimization, and compilation for the DPU. Experimental results demonstrate that the proposed system outperforms comparable processing speed and accuracy implementations while maintaining energy efficiency. The study highlights the viability of deploying lightweight deep learning models on FPGA platforms for real-time embedded vision applications in intelligent transportation systems.

Heller et al. [13] explore the deployment of YOLO-based deep learning models for real-time marine object detection on embedded edge devices. The authors train and optimize YOLOv4 and YOLOv4-Tiny models using techniques like structured pruning, quantization (INT8 and FP16), and architecture modifications to improve inference speed while maintaining accuracy. These models have been deployed on Nvidia Jetson Xavier AGX, AMD-Xilinx Kria KV260, and Intel Movidius Myriad X VPU. Results show that YOLOv4-Tiny on Jetson Xavier achieved up to 90 FPS with minimal accuracy loss, while the Movidius VPU offered the best performance-per-watt efficiency. The Kria KV260 provided a strong balance of speed and energy efficiency, making it suitable for low-power applications. The study demonstrates the feasibility of efficient, high-speed object detection on resource-constrained devices for marine environments.

Montgomerie-Corcoran et al. [14] introduce SATAY, a tool flow for automatically generating high-performance FPGA-based accelerators for YOLO object detection models, supporting versions from YOLOv3 to YOLOv8. Unlike previous approaches that rely on heterogeneous architectures, SATAY implements fully on-chip, deeply pipelined streaming designs that minimize off-chip communication and maximize parallelism. It incorporates novel hardware components for YOLO-specific operations, applies quantization and buffer optimization to reduce resource usage, and performs automated design space exploration. The evaluation shows that SATAY-generated designs outperform existing FPGA accelerators and achieve competitive energy efficiency compared to embedded GPUs and CPUs, marking the first reported YOLOv8 implementation on FPGAs.

Nguyen et al. [15] propose an FPGA-SoC implementation of the YOLOv4 object detection model for real-time flying-object detection, particularly UAVs and aircraft. The authors adapt and optimize YOLOv4 and YOLOv4-Tiny for deployment on the Xilinx ZCU104 FPGA platform using Vitis AI. Key contributions include network pruning, quantization, and architectural modifications to accommodate FPGA constraints, such as replacing unsupported layers and adding a YOLO detection layer for better small-object detection. The pruned YOLOv4 model achieves performance close to the original while significantly reducing model size and improving frame rate (29 FPS). The implementation demonstrates superior power efficiency—2–3× over RTX 2080Ti and 3–4× over GTX 1650—making it suitable for embedded applications requiring real-time processing and low energy consumption.

Valadanzoj et al. [16] propose a novel FPGA-based accelerator for the YOLOv4-Tiny object detection model, targeting real-time performance in self-driving automotive applications. The architecture leverages low-precision quantization (8-bit for data and 5-bit for weights) and introduces a double-multiplication technique using a single DSP block to increase parallelism and reduce resource usage. A Genetic Algorithm (GA) optimizes layer-specific quantization parameters, maintaining model accuracy. Implemented on a Xilinx Zynq ZC706 FPGA, the design achieves 55 FPS at 79% mAP, representing a 13% speed increase over prior work with only a 3% drop in accuracy. The system also demonstrates reduced DSP and memory usage, enhancing cost-effectiveness for embedded automotive applications.

III. IMPLEMENTATION

Implementing the proposed object detection and classification system for edge FPGAs comprises several structured steps. Figure 1 illustrates the overall system architecture.

*A. Dataset Preparation*

The proposed system utilizes two widely recognized datasets to support general-purpose and application-specific object detection and classification tasks. The COCO dataset [17] is a large-scale benchmark dataset for object detection, segmentation, and captioning. It contains over 200,000 labeled images with over 80 object categories, capturing various real-world scenarios with varying scales, occlusions, and lighting conditions. COCO is particularly valuable for training deep learning models to perform robust detection in complex environments. In contrast, the GTSRB dataset [18] focuses on a specific application domain—traffic sign recognition. It comprises over 50,000 images of traffic signs categorized into 43 different classes, collected under real-world driving conditions. The dataset presents challenges such as motion blur, varying illumination, and partial occlusions, making it ideal for evaluating the system's performance in intelligent transportation applications. By employing the COCO and GTSRB datasets, the proposed system is trained to achieve high accuracy in general object detection and domain-specific classification tasks, enhancing its versatility and applicability in edge scenarios.

*B. Model Training*

This study selects YOLOv5 for the proposed object detection and classification system due to its favorable trade-off between detection accuracy and computational efficiency, making it suitable for deployment on resource-constrained edge devices. The model training has been conducted using the PyTorch framework, with YOLOv5 initialized using pre-trained weights to accelerate convergence and enhance performance. The training process involved fine-tuning the model on two benchmark datasets: COCO for general object detection and GTSRB for traffic sign classification. Appropriate data augmentation techniques were applied during training to improve the model's ability to generalize to diverse environments. The training configuration included a learning rate of 0.01, batch size of 32, and 100 training epochs, using the stochastic gradient descent (SGD) optimizer with a momentum of 0.937 and weight decay of 0.0005. These hyperparameters were empirically selected to balance training stability, convergence speed, and final model accuracy. Early stopping and model checkpointing strategies were also employed to prevent overfitting and ensure the retention of the best-performing model weights. This training process effectively prepared YOLOv5 for integration into the proposed FPGA-based real-time object detection and classification system.

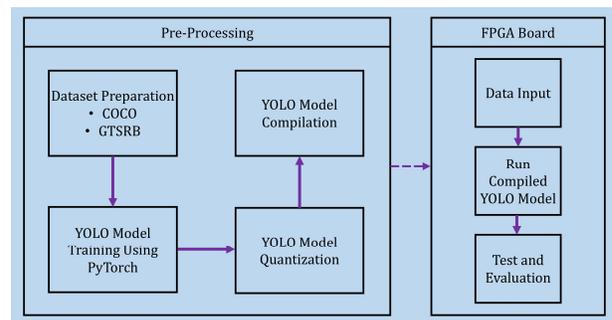

Fig. 1. Overall architecture of the proposed object detection and classification system.

*C. Model Quantization and Compilation*

A post-training quantization and compilation workflow was applied using the Vitis AI development environment to enable efficient deployment of the trained YOLOv5 model on the Xilinx Kria KV260 FPGA. The quantization process involved converting the model's 32-bit floating-point weights and activations into 8-bit integer (INT8) representations using the Vitis AI toolchain. This step significantly reduced the model's memory footprint and improved data transfer efficiency. It lowered computational complexity, all of which are essential for meeting the strict resource and power constraints of edge FPGAs. Quantization-aware techniques ensured that the reduction in numerical precision introduced minimal impact on the model's detection accuracy. The YOLO model has been compiled following quantization into a hardware-friendly format specifically optimized for execution on the Deep Processing Unit (DPU) of the KV260 FPGA. This compilation step involved graph-level optimizations, operator mapping, and memory allocation tailored to the architecture of the target platform. The combined quantization and compilation process allowed the YOLOv5 model to achieve real-time inference performance with low latency and power consumption, demonstrating its suitability for object detection and classification tasks in edge computing applications.

## IV. RESULTS AND DISCUSSION

The optimized YOLOv5 model was deployed onto the Xilinx Kria KV260 FPGA board to enable real-time inference on an edge device following the quantization and compilation stages. The model execution was carried out on the board's integrated DPU, which is specifically designed to accelerate deep learning workloads. Input images were supplied to the board, and the system performed object detection and classification directly on the edge device without reliance on external processing. The deployed model was then evaluated to assess its performance in inference latency, resource utilization, and operational efficiency under real-world scenarios. The results demonstrated that the edge deployment maintained consistent detection accuracy while meeting the latency constraints required for real-time applications.

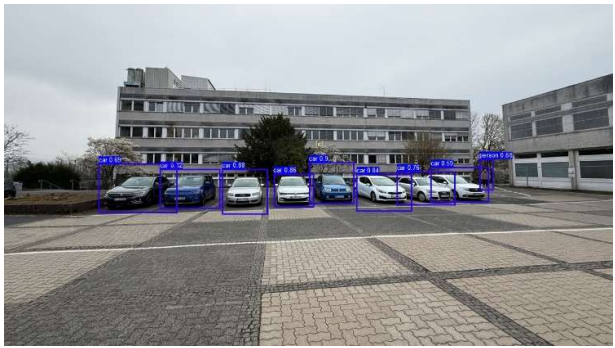

Fig. 2. Object detection and classification result using COCO dataset.

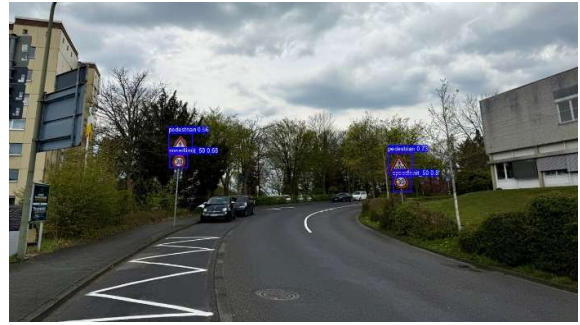

Fig. 3. Object detection and classification result using GTSRB dataset.

*A. Experimental Results*

The proposed object detection and classification system has been evaluated using two standard benchmark datasets: COCO for general object detection and classification and GTSRB for traffic sign detection and classification. These datasets were selected to validate the system's effectiveness across generic and application-specific tasks. A total of 72 test images were curated to represent diverse real-world scenarios, including variations in lighting conditions, object scales, background complexity, and occlusions. These images were then processed by the Xilinx Kria KV260 FPGA board, where the optimized and quantized YOLOv5 model was deployed for inference using the board's integrated DPU.

The primary objective of the evaluation was to assess the feasibility and efficiency of executing the object detection and classification system on a resource-constrained edge FPGA platform. The deployed system achieved a classification accuracy of 98.6% while maintaining an inference speed of 9 FPS, demonstrating its capability to perform real-time processing under edge deployment constraints. Fig. 2 presents representative output samples from the object detection and classification task using the COCO dataset. In addition, Fig. 3 illustrates output examples for traffic sign detection and classification using the GTSRB dataset. These results highlight the system's potential for accurate and efficient deployment in real-world edge computing environments.

*B. Comparison with Related Works*

Table 1 presents a comparative analysis of various FPGA-based object detection systems using different YOLO model variants across multiple studies. While prior works predominantly utilized YOLOv3 or YOLOv4 Tiny models on FPGA platforms such as ZC706, ZCU104, and Kria KV260, the proposed work employs YOLOv5 on the Kria KV260 board. In terms of classification accuracy, the proposed system achieves 98.6%, which is competitive with the highest reported value of 99% by Amin et al. [19] and significantly higher than the rest. Although the inference speed (9 FPS) is lower than other implementations—particularly Nguyen et al. [15], which achieved 125 FPS—it reflects a trade-off between speed and accuracy. The proposed system maintains a low power consumption of 3.5 W, comparable to Amin et al. [19], and favorable for edge deployment. In terms of efficiency measured as frames per second per watt (FPS/W), the system achieves 2.6 FPS/W, indicating balanced performance with a strong emphasis on accuracy and power efficiency for edge computing scenarios. Despite the slightly reduced inference speed, the system's better accuracy and power efficiency make it a viable solution for resource-constrained edge devices.

TABLE I. COMPARISON OF THE PROPOSED SYSTEM WITH OTHER RELATED WORKS.

| Evaluation Metrics / Author | Valadanzoj et al. [16] | Nguyen et al. [15] | Heller et al. [13] | Amin et al. [19] | Ngo et al. [12] | This Work |
|---|---|---|---|---|---|---|
| Model | YOLO v4 Tiny | YOLO v4 Tiny | YOLO v4 Tiny | YOLO v3 Tiny | YOLO v3 Tiny | YOLO v5 |
| FPGA Board | ZC 706 | ZCU 104 | Kria KV 260 | Kria KV 260 | Kria KV 260 | Kria KV 260 |
| Accuracy (%) | 79 | 78 | 75 | 99 | 91.87 | 98.6 |
| Inference Speed (FPS) | 55 | 125 | 15 | 15 | 25 | 9 |
| Power (W) | 13.6 | 26.4 | 8 | 3.5 | 7.9 | 3.5 |
| Efficiency (FPS/W) | 4 | 4.7 | 1.87 | 4.2 | 3.1 | 2.6 |

## V. CONCLUSION AND FUTURE WORK

This study introduced a resource-efficient, real-time object detection and classification system based on the YOLOv5 model, optimized for deployment on edge FPGA platforms. The system has been successfully implemented on the Xilinx Kria KV260 board by leveraging model quantization and compilation techniques, achieving a balance between accuracy, power consumption, and real-time performance. Evaluation of standard datasets demonstrated the system's effectiveness for general and application-specific tasks. Compared to existing works, the proposed approach offers improved accuracy and energy efficiency, making it well-suited for edge applications such as intelligent transportation and embedded vision. Future studies will focus on improving the inference speed of the system while preserving accuracy and resource efficiency. Additionally, the model will be extended to support multi-class detection tasks and evaluated under more diverse real-world deployment scenarios.

ACKNOWLEDGMENT

This research work is part of the IGNITE project, financially supported by the Freiraum program of Stiftung Innovation in der Hochschullehre.